\pdfoutput=1
\documentclass[11pt]{article}
\usepackage{graphicx}
\usepackage{listings}
\usepackage[dvipsnames]{xcolor}
\usepackage{color}
\definecolor{deepblue}{rgb}{0,0,0.5}
\definecolor{deepred}{rgb}{0.65,0,0}
\definecolor{deepgreen}{rgb}{0,0.5,0}
\usepackage[]{acl2023}

\usepackage{times}
\usepackage{latexsym}

\usepackage[T1]{fontenc}
\usepackage[utf8]{inputenc}

\usepackage{microtype}
\usepackage{mathtools}

\title{\textsc{CoRRPUS}: Code-based Structured Prompting for Neurosymbolic Story Understanding}

\author{Yijiang River Dong$^1$, Lara J. Martin$^2$\thanks{~~Work done while at the University of Pennsylvania.} , \and Chris Callison-Burch$^1$\\
  $^1$University of Pennsylvania\\
  $^2$University of Maryland, Baltimore County\\
  \texttt{riverd@sas.upenn.edu, laramar@umbc.edu, ccb@seas.upenn.edu } 
  }

\lstdefinestyle{mystyle}{language=Python, 
basicstyle=\fontsize{6}{8}\selectfont\ttfamily, 
breaklines=true, 
keywordstyle=\bfseries\color{deepblue}, 
morekeywords={},
emph={self}, 
emphstyle=\bfseries\color{deepred}, 
commentstyle=\itshape\color{black!50!white}, 
stringstyle=\bfseries\color{deepgreen},
frame=single,
showstringspaces=false,
tabsize=2,
captionpos=b}
\begin{document}

\maketitle
\begin{abstract}
Story generation and understanding---as with all NLG/NLU tasks---has seen a surge in neurosymbolic work. Researchers have recognized that, while large language models (LLMs) have tremendous utility, they can be augmented with symbolic means to be even better and to make up for many flaws that neural networks have.
However, symbolic methods are extremely costly in terms of the amount of time and expertise needed to create them.
In this work, we capitalize on state-of-the-art Code-LLMs, such as Codex, to bootstrap the use of symbolic methods for tracking the state of stories and aiding in story understanding.
We show that our CoRRPUS system and abstracted prompting procedures can beat current state-of-the-art structured LLM techniques on pre-existing story understanding tasks (bAbI Task 2 and Re$^3$) with minimal hand engineering.
This work highlights the usefulness of code-based symbolic representations for enabling LLMs to better perform story reasoning tasks.
\end{abstract}

\section{Introduction}

Stories are a complex form of writing that involve many inter-dependent components, such as causal reasoning \cite{Mostafazadeh2020,Han2021}, temporal ordering \cite{Basu2021}, social commonsense \cite{Hwang2021}, and the consistent portrayal of facts and events as the story unfolds \cite{Yu2021,Wilmot2021}.
Despite the recent uptick in research involving story understanding and reasoning \cite{Qin2019, Han2019, Mostafazadeh2020, Peng2021, brahman-etal-2021-characters-tell, Martin2021, Brahman2022, Chen2022, Li2022, Guan2022, Andrus2022}, there is still a significant gap in the abilities of models to actually understand (or generate) stories at a reasonable level of commonsense. \citet{Ranade2022} provide a recent overview of the area.

\begin{figure}[h!]
\includegraphics[width=0.48\textwidth]{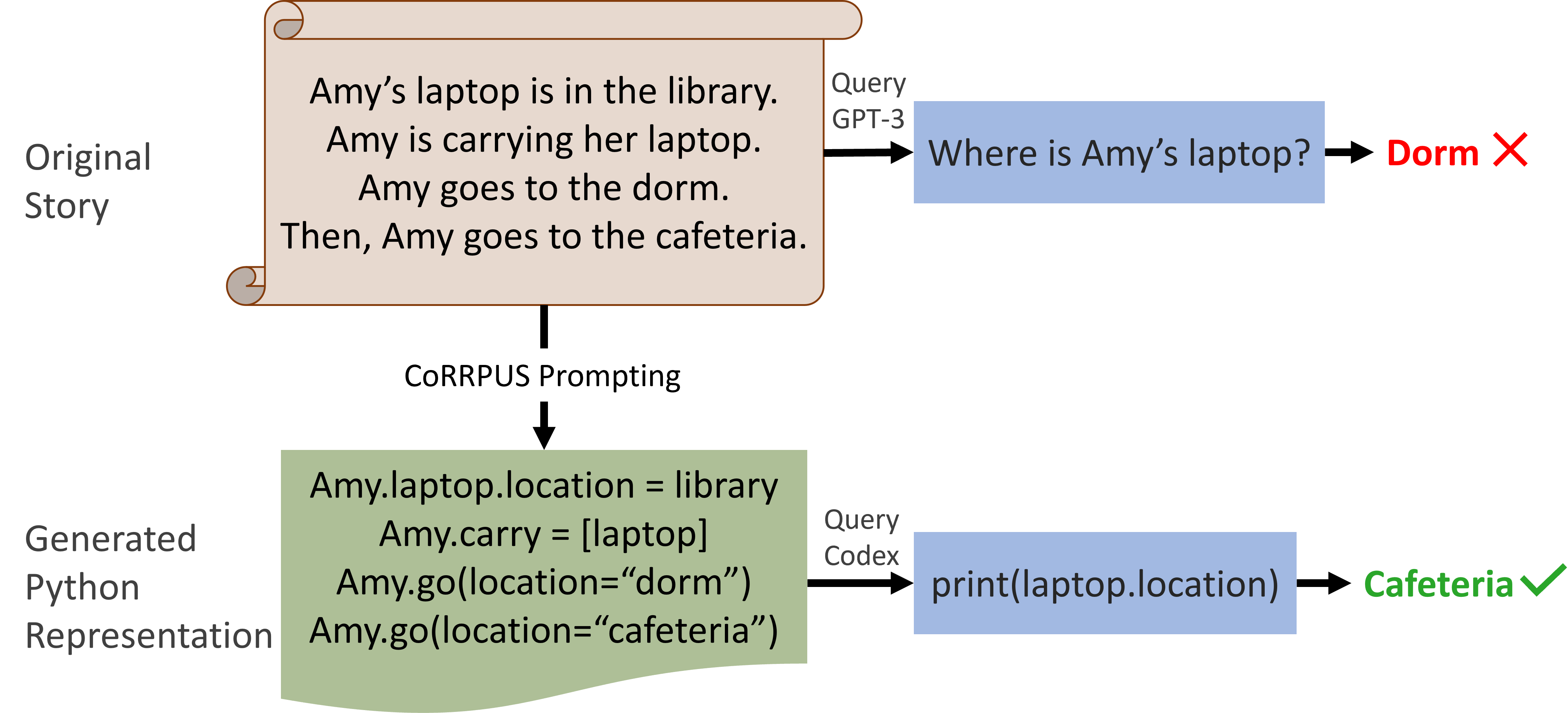} 
\caption{CoRRPUS prompting aids the LLM in properly following the laptop throughout the short story. Meanwhile with natural language prompts, GPT-3 fails to keep track of the laptop through the character Amy's movement. It can follow that Amy has her laptop and that she brought it to the dorm, but it ``loses track'' of the laptop after Amy goes to the cafeteria.}
\label{fig:intro}
\end{figure}

Although large language models (LLMs) like GPT \cite{GPT2, brown2020language} by themselves can produce text that can be indistinguishable from human-written stories, these models still struggle to generate coherent long-form text \cite{See2019} \& solve simple commonsense reasoning tasks (See Figure \ref{fig:intro} for an example.) and therefore, end up only writing at human-level quality less than three-quarters of the time \cite{Ippolito2020}. It is also likely that the majority of this generated text is actually memorized and generated verbatim from human-written stories \cite{Lee2022}.

Over the past few years, researchers have begun to see value in integrating symbolic AI methods---that are consistent and logical---with neural networks---that are flexible and unpredictable \cite{Garcez2020}.
These neurosymbolic techniques try to balance the best of both worlds.
\citet{nye2021improving} draws a parallel between neurosymbolic AI and the System 1/System 2 cognitive science theory \cite{evans2003two}: neural sequence models are a fast and intuitive system (System 1) that can be improved by adding a logical reasoning system (System 2).
\citet{Martin2021} and \citet{nye2021improving} use GPT-2/3, respectively, to produce candidate story continuations and then use a symbolic state representation to compare against in order to remove suggestions that would be inconsistent.

We extend this work with the use of the Code-LLM Codex \cite{chen2021evaluating} for structuring and tracking state changes for story understanding. Our system, which we call \textsc{CoRRPUS}
({\bf Co}de \textbf{R}epresentations to {\bf R}eason \& {\bf P}rompt over for {\bf U}nderstanding in {\bf S}tories)\footnote{Code can be found here: \url{https://github.com/dong-river/CoRRPUS}}, 
extracts structured information (the story world model) which is then used for reasoning. We compare to neural-only, symbolic-only, and prior works' neurosymbolic systems on two pre-existing story understanding tasks: bAbI and Re$^3$, which we will describe in Section \ref{sec:experiments}.

Our contributions are as follows: 

1. We adapt the Code-LLM model for modeling story worlds and tracking information over time. Our CoRRPUS technique outperforms existing state-of-the-art models on bAbI and Re$^3$ story understanding tasks.

2. We explore various prompting styles to achieve the best performance from Code-LLMs and report on best practices for working with Code-LLMs.

%3. We conduct extensive experiments to show that by performing neurosymbolic reasoning, CoRRPUS outperforms many previous methods in bAbI reasoning tasks and Re$^3$ story inconsistency detection task.

In the rest of the paper, we will outline a brief history of neurosymbolic work in storytelling and recent use of Code-LLMs. We will then go over CoRRPUS's three prompting methods, followed by a description of the two story understanding tasks bAbI and Re$^3$ and our experiments with CoRRPUS on both.

\section{Related Work}
\subsection{Neurosymbolic Reasoning in Storytelling}

Structured representations have existed in story generation for decades (e.g., \cite{Lebowitz1984,Turner1985}), and while pure symbolic methods are still researched today \cite{Garbe2019,Christensen2020,Ware2021}, a recent push for combining these structured symbolic methods with probabilistic neural networks has grown. We outline some of these methods here.

\begin{figure*}[t!]
\includegraphics[scale=0.8]{figures/babi_example.pdf}
\caption{An example prompt used for bAbI Task 2. All three prompting methods have the same prompt initialization (a) followed by their respective additional functions (b, c, or d), found inside the World class (end of a). That is, for the 1-shot example, CoRRPUS would be provided (a) + (b, c, or d) depending on the prompting method. To prompt for the next story, CoRRPUS is given (a) + the non-highlighted of (b, c, or d). The highlighted section would then be generated by CoRRPUS.}
\label{prompts}
\end{figure*}

Much of the research in story generation uses a hierarchical or multi-stage technique to first plan out the underlying plot (also known as \textit{fabula}) or other underlying structures, and then generate natural language sentences that are informed by the structure \cite{Martin2018,fan2018hierarchical,Tambwekar2019,Yao2019,Goldfarb-Tarrant2019,Ammanabrolu2020,rashkin2020plotmachines,guan2020knowledge,sun-etal-2022-summarize,yang2022re3}.

Other popular ways to include structure to enhance story generation include using external resources (\citet{Huang2020} with ConceptNet \cite{ConceptNet}, \citet{Martin2021} with VerbNet \cite{VerbNet}, \citet{Peng2021} with ATOMIC \cite{ATOMIC}) or by extracting information from the original CoRRPUS beyond the series of events, such as characters' emotions toward each other \cite{RaoVijjini2022} or knowledge triplets \cite{Alabdulkarim2021}.
Similarly, summary information of the story so far \cite{Callison-Burch2022}, and a combination of summaries and character relationships \cite{si-etal-2021-telling} have been used to augment storytelling dialog generation.

Researchers have also seen the benefit of neurosymbolic methods by using external knowledge bases \cite{Zhang2021} or symbolic representations \cite{Li2022} to augment a transformer's ability to perform story understanding.

Most relevant to our work, however, is the extraction of a story world or world model.
This work uses a dynamic external representation in order to keep track of the story while the system is generating in order to make for more consistent stories. So far this has been done with world state dictionaries \cite{Martin2021}, object-oriented notation \cite{nye2021improving}, and knowledge graphs \cite{Andrus2022}.

\subsection{Introduction of Code-LLMs for Reasoning}
Recently, thanks to the LLMs and large-scale code training data \cite{husain2019codesearchnet}, many breakthroughs has been made in automatic code synthesis \cite{feng2020codebert,clement2020pymt5,chen2021evaluating}. Code-LLMs have been shown to be adept at logical reasoning \cite{wei2022chain}, numerical reasoning \cite{cobbe2021training}, theorem proving \cite{wu2022autoformalization}, and linguistic reasoning tasks, such as the command composition task SCAN \cite{lake2018generalization} as shown by \citet{zhou2022least}.

\citet{madaan2022language} show that Code-LLMs perform better than LLMs in various structured commonsense reasoning tasks including procedural reasoning and entity state tracking, where they used a graph-based representation. 
Similarly, we prompt Code-LLMs to extract structured world model for neurosymbolic story understanding.

\section{The CoRRPUS Prompting System}
\label{sec:corrpus}
In this work, we examine the types of prompts for tracking symbolic story state representations using OpenAI's Code-LLM, Codex \cite{chen2021evaluating}. 
All experiments that use the CoRRPUS prompt system are conducted with code-davinci-002 using OpenAI's API (which was free for research purposes). Following \citet{holtzman2019curious}, we use nucleus sampling with top-p value equals to 0.95. We set the temperature to 0 when no diverse generation is needed (i.e., answering multiple choice questions from the bAbI task, Section \ref{sec:babi}) and set the temperature to 0.7 when diverse generation is needed (such as in the Re$^3$ task, Section \ref{sec:re3}). All the experimental results come from a single run.

We provide the Codex model a collection of classes in Python to represent a model of characters and objects that we want to track in the story, in addition to an initialization of the specific characters and objects for the current story, which is presented in a \texttt{World} class. See Figure \ref{prompts}a and Appendix \ref{re3_examples} for examples of these classes. In addition to this information for the current story, we also provide Codex a 1-shot example of the full process we want it to complete.

Formally, we define the problem as: given a story $\mathcal{S}=[S_0: S_1: ...: S_n]$ and a story world state initialization $\mathcal{W}_0$, we want the model to update the story world state $\mathcal{W}_i$ until the story is complete $\mathcal{W}_n$.
After each sentence $\mathcal{S}_i$ of the story, the model updates the world state using some black-boxed update function $\mathcal{U}_i$. Thus, the final world state $\mathcal{W}_{n}$ is obtained via $\mathcal{W}_0 \xrightarrow[]{\mathcal{U}_0} \mathcal{W}_1  \xrightarrow[]{\mathcal{U}_1} .... \xrightarrow[]{\mathcal{U}_{n-1}} \mathcal{W}_{n-1}  \xrightarrow[]{\mathcal{U}_{n}} \mathcal{W}_{n}$.
Note that, with the except of $\mathcal{W}_0$ and $\mathcal{W}_n$, these intermediate world states are also opaque and unseen from the user's perspective.
This system, which we call CoRRPUS, extracts structured information using code-based chain-of-thought--type prompting in order to more accurately track the underlying story state and detect any inconsistencies. 

We experiment with three different prompting techniques. The following examples show how the story sentence ``Sandra journeyed to the bedroom'' would be modified for each prompt type.
\begin{itemize}
\item \textbf{Comment Only:} This is the simplest prompt, which is given no extra structural information. Thus, it has to rely directly on the comments from the prompt and the story world state initialization. This shows us how well the Code-LLM can infer and fill in information with very little guidance. Example: Given the comment \texttt{\#\# Sandra journeyed to the bedroom.}, the model should generate \texttt{self.Sandra.location = "bedroom"}.
\item \textbf{Specific Functions:} This prompt converts each individual sentence into a specific function. In addition to the commented sentences at the beginning of the prompt, the system is also prompted to generate the functions for each sentence. Example: Given the function name \texttt{Sandra\_journeyed\_to\_the\_
bedroom()}, the model should generate a definition for the function, which should contain \texttt{self.Sandra.location = "bedroom"}.
\item \textbf{Abstract Functions:} This last prompting style provides the model with functions for actions or setting attributes. The model then does not have to figure out what it means when a particular event happens but still has to map which function is appropriate for a given story sentence and how to fill in the arguments of the function.   Example: Given \texttt{go(character, location)} and other functions, generate \texttt{go(character=Sandra, destination=bedroom)}.

\end{itemize}
Full examples of these prompts can be shown in Figure \ref{prompts} and Appendix \ref{re3_examples}.

\begin{table*}[ht!]
\centering
\begin{tabular}{l|ccc|cc}

& \multicolumn{3}{c|}{\textbf{Original Results}} & \multicolumn{2}{c}{\textbf{Our Results (1-shot)}}\\
\textbf{Method}    &\textit{Model} & \textit{\# Shots}& \textit{Accuracy} &  \textit{Model}&\textit{Accuracy}  \\ \hline\hline
Random & -& - & - & N/A & 25\%\\\hline
GPT-3 \cite{nye2021improving} & GPT-3 & 0                      & 29.0\% & GPT-3 &56.5\%\\
Codex w/ natural language & - & -& - & Codex & 57.8\%\\\hline
COT \cite{creswell2022selection}& 7B LLM & 5  & $\sim$30\%& GPT-3 &  46.4\% \\ 
COT  \cite{creswell2022selection} & 280B LLM& 5 & $\sim$35\%  & GPT-3 & 46.4\% \\ 
SI  \cite{creswell2022selection}& 7B LLM & 5  & $\sim$30\%& GPT-3 &  29.3\%\\
SI  \cite{creswell2022selection} & 280B LLM& 5  & Not reported & GPT-3 & 29.3\%\\ 
DS \cite{nye2021improving}  & GPT-3&  10 &\textbf{100.0\%}& -&-\\ \hline
\textbf{CoRRPUS (comment only)}        & -     &-    &-&  Codex & 67.0\%    \\
\textbf{CoRRPUS (specific functions)}& -         &-   &    -  & Codex &78.7\%  \\
\textbf{CoRRPUS (abstract functions)} & -          &-  &   -  & Codex &\textbf{99.1\% }\\

\end{tabular}
\caption{Accuracy on bAbI task 2. We report other systems' accuracy with the number of examples (\# shots) used for their prompts from their respective papers. Chain-of-Thought (COT) and Selection-Inference (SI) prompting are by \citet{creswell2022selection} and the Dual-System (DS) is by \citet{nye2021improving} using a 5-shot prompt.
COT \& SI numbers are approximations since they were reported in graph format (Figure 4b of \citet{creswell2022selection}).
All systems were reimplemented using GPT-3 (175B parameters), with 1-shot prompting to match our experiments.
}
\label{bAbI_results}
\end{table*}

\section{Experiments}
\label{sec:experiments}
To show how CoRRPUS can improve story understanding via maintaining a world model, we evaluate our system on two tasks: (1) Task 2 of the bAbI set of tasks \cite{bAbI}---which tests multi-step reasoning, and (2) Re$^3$ \cite{yang2022re3}, for detecting story inconsistencies in complicated real-world story examples. 

\subsection{Introduction to the Story Understanding Tasks}
\textbf{bAbI} \cite{bAbI} is a set of tasks on simple stories that ask questions about what happened during the story. The tasks have various ways of responding to the questions such as with argument relations, supplying supporting acts, or a simple yes/no. Following \citet{nye2021improving}, we use only bAbI Task 2, which focuses on questions tracking characters carrying objects and moving between different locations. \citet{creswell2022selection} also look at Tasks 1, 3, 15, \& 16. Since Tasks 1-3 are the same except for differences in story length and Tasks 15 \& 16 are much easier tasks, as we found by \citet{creswell2022selection}'s results, we decided to focus on Task 2.

\textbf{Re$^3$} \cite{yang2022re3} is a story inconsistency detection task aimed at identifying character-based contradictions in stories. Similar to \citet{Qin2019}, \citet{yang2022re3} built a set of stories, each with a variation that is counterfactual to the original story. These stories were generated by LLMs to generate consistent/contradictory stories based on a story premise and then the model was asked to detect any story inconsistencies. \citet{yang2022re3} make use of the Edit function for GPT-3 to correct the detected factual inconsistencies in order to maintain long-range story consistency.

These two tasks have been seen to be extremely challenging for LLMs with naive, few-shot prompting leading to accuracy barely above random chance.

\subsection{Question-and-Answering (bAbI Task 2)}
\label{sec:babi}
bAbI \cite{bAbI} is a question answering dataset for logic-based reasoning tasks. In Task 2, it first provide a story S focusing on the movement of objects and characters throughout the story and a query Q about their locations. The dataset contains 1,000 testing samples of (S, Q, A) tuples, where A is the answer to the query. We choose Task 2 because of the recent neurosymbolic reasoning work evaluated on it \cite{nye2021improving,creswell2022selection}, which we use as our baselines.

\paragraph{The CoRRPUS Formulation for bAbI.} 
Starting with the CoRRPUS prompt formulation that was described in Section \ref{sec:corrpus}, we first initialize the \texttt{character} class with the attributes \texttt{name}, \texttt{location}, and \texttt{inventory}. The object class has attributes \texttt{name}, \texttt{location}, and \texttt{carrier}. Then, we let Codex complete the \texttt{World} class by generating the rest of the \texttt{story()} function, which tracks the story state changes and ends with a \texttt{print()} function that gives the answer to the query Q given in the bAbI Task. We then measure the accuracy of the model selecting the correct answer.

We compare our CoRRPUS system to the following baselines for bAbI Task 2:
\begin{enumerate}
\item  \textbf{Random:} This is just the likelihood of selecting the right multiple choice answer randomly.
\item  \textbf{GPT-3:} We compare an off-the-shelf GPT-3 davinci model using one-shot prompting to the results from \citet{nye2021improving} who use zero-shot prompting.
\item \textbf{Codex with Natural Language Prompt:} These are the same natural language prompts as the \textbf{GPT-3} baseline but using the Codex model instead. This method will highlight any performance boost from using the Codex model without our CoRRPUS prompts.
\item  \textbf{LLM with Chain-of-Thought (COT) Prompting \cite{creswell2022selection}:} Chain-of-Thought Prompting \cite{wei2022chain} is the process of prompting an LLM to include reasoning traces for solving a given task, and it has been shown to improve model performance on various tasks. 
\citet{creswell2022selection}'s models were a 7 billion- and 208 billion-parameter LLMs from the Gopher family \cite{rae2021scaling}, and they used 5-shot prompting.
In addition to reporting their original accuracy results from their models, we also rerun their experiments using GPT-3 (175 billion parameters) using one-shot prompting to match our experiments.
\item  \textbf{LLM with Selection-Inference (SI) \cite{creswell2022selection}:} This method prompts the LLM to first select sentences relevant to the question, reveal the reasoning chain, and then do the inference. Again, \citet{creswell2022selection} use five-shot prompting with the 7 billion- and 280 billion-parameter models, and we reimplemented the Selection-Inference framework using GPT-3 and one-shot prompting. 
\item  \textbf{GPT-3 Dual-System (DS) \cite{nye2021improving}:} This method is based off of System 1/System 2 thinking \cite{evans2003two}. They specify the functions of all actions used in bAbI Task 2 to create a logical, symbolic component (System 2). Then the system prompts GPT-3 to match each sentence with the actions (System 1) and executes the corresponding function. We did not rerun the results from this experiment.
\end{enumerate}

\begin{figure*}
\centering
\includegraphics[scale=0.4]{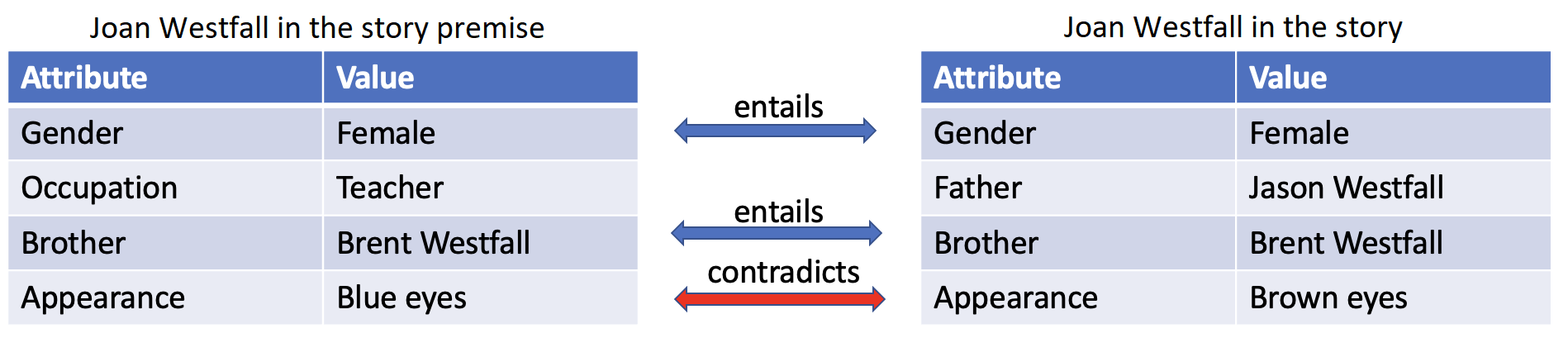}
\caption{An illustration of the contraction detection process for Re$^3$. Following \citet{yang2022re3}, we check attributes across the premise and the story using a BART-Large-based entailment model and flag any contradictions for attribute-value pairs with the same attribute key. In this example, the ``appearance'' attribute would be flagged as a contradiction.}
\label{re3_check}
\end{figure*}

\paragraph{Results and Discussion.} As show in Table \ref{bAbI_results}, even the one-shot prompting CoRRPUS system with Comment Only or Specific Functions achieve 12\% and 22\% higher accuracy, respectively, compared to vanilla one-shot GPT-3. We believe that is because GPT-3 when used out-of-the-box and unaided is known to be bad at multi-step reasoning \cite{Shridhar2022,Sap2022}.  Meanwhile, Code-LLMs represent knowledge symbolically and therefore are better at logical reasoning \cite{madaan2022language}.

However, the increase in accuracy is not solely due to Codex. When we use the same natural language prompt as the GPT-3 one-shot experiment, the accuracy only goes up 1.3\%. It isn't until we introduce the CoRRPUS structured prompting system that we start to see accuracies between 67\%-99.1\%.
We see that when the underlying functions for actions are provided for CoRRPUS (Abstract Functions), it can achieve near-perfect accuracy (99.1\%), vastly exceeding other prompting methods. 
This experiment shows the importance of abstraction to trackability and composability in symbolic reasoning.
The other system which approximately matches CoRRPUS in accuracy is the GPT-3 Dual System \cite{nye2021improving}, which splits the reasoning steps into two separate systems, uses highly-detailed hand-written rules, and requires 10-shot prompting. We show that providing a simpler one-shot prompt to Codex is enough to perform logic reasoning on these simple stories, as long as it is provided low-level functions to compute over.

With such high accuracies from our CoRRPUS system and \citet{nye2021improving}'s Dual System, we are tempted to consider bAbI's Task 2 a solved problem.

\subsection{Detecting Story Inconsistencies (Re$^3$)}
\label{sec:re3}
Given the simplicity of the sentences in bAbI, we wondered how well CoRRPUS could process and understand stories with complicated real-world sentences, such as those in the Re$^3$ task.
This dataset contains 50 (P, P', S, S') tuples where P denotes a story premise, P' is a premise contradictory to P, S is the story generated from P, and S' is the story generated following P'. The task is framed as classification; one wants to flag (P, S) and (P', S') as consistent and (P, S') and (P', S) as contradictory. They then report the ROC-AUC (area under the ROC curve) score. An example of the comparison between facts of the premise and the story is shown in Figure \ref{re3_check}.

\paragraph{CoRRPUS Formulation for Re$^3$.} Following our CoRRPUS prompt starting point (see Section \ref{sec:corrpus}), we initialize the \texttt{character} class with common person attributes including \texttt{name}, \texttt{appearance}, \texttt{occupation}, \texttt{gender}, \texttt{age}, and \texttt{relations} (to other characters). Then we initialize the main characters in the \texttt{World} class and one-shot prompt Codex to complete the \texttt{story()} function for tracking the state changes of the characters. Full examples of each of these prompts can be found in Appendix \ref{re3_examples}.
Once the \texttt{World} state is complete, it is fed into GPT-3 to be converted into natural language text. We then, following the methods of \citet{yang2022re3}, pass this natural language story state to BART-Large to find any contradictions between the story and the original corresponding premise from the dataset.

\noindent We use the following baselines for the Re$^3$ experiment:
\begin{enumerate}
\item \textbf{GPT-3:} We query GPT-3 using zero-shot prompting to determine whether there are inconsistencies in the (S, P) pair. 
\item \textbf{Textual Entailment \cite{yang2022re3}:} This method uses the BART-Large-based \cite{lewis2020bart} entailment model trained on MNLI \cite{MNLI} to score (S, P) pairs.
\item  \textbf{Entailment-DPR \cite{yang2022re3}:} For each sentence $s_i$ in S, this method computes its relevance score against each sentence in P by using Dense Passage Retrieval \cite{karpukhin2020dense}. Then the method takes the sentence with highest relevance score $p_i$ and use the entailment model to detect contradictions.
\item \textbf{Structured-Detect \cite{yang2022re3}: }This method prompts GPT-3 to extract an attribute dictionary for each character in the story premise and the story. To prevent hallucinations, the method prompts GPT-3 three times and then uses the BART-Large-based entailment model to take the most-agreed attributes. Finally, the method converts the attribute-value pairs into natural sentences and uses the entailment model to detect contradictions for values under the same key.
\end{enumerate}

Inspired by the Structured-Detect \cite{yang2022re3} and Self-Consistency \cite{wang2022self} methods, we ask CoRRPUS to complete 3 different generations with temperature equal to 0.7 and select the attribute-value pair by majority voting. Finally, like \citet{yang2022re3}, we use a BART-Large-based entailment model to flag any contradictions for the values of any attributes found in both the premise and the story. A toy example of this comparison process is shown in Figure \ref{re3_check}.

\begin{table}[h]
\centering
\begin{tabular}{lc}
\hline
\textbf{Method}  & \textbf{ROC-AUC} \\ \hline\hline
Random & 0.5 \\
GPT-3            & 0.52             \\\hline
\textit{\citet{yang2022re3} baselines:}&\\
Entailment    & 0.528             \\
Entailment-DPR   & 0.610             \\
Structured-Detect & 0.684             \\ \hline
\textbf{CoRRPUS (comment only) }           &     0.751             \\
\textbf{CoRRPUS (specific functions)}            &  0.794               \\
\textbf{CoRRPUS (abstract functions)   }         &      0.704           \\
\end{tabular}
\caption{ROC-AUC score on the Re$^3$ consistency detection task. The scores for Entailment, Entailment-DPR, and Structured-Detect models are directly cited from \citet{yang2022re3}.}
\label{tab:Re3}
\end{table}

\paragraph{Results and Discussion.} Table \ref{tab:Re3} shows that all version of CoRRPUS greatly outperform the baseline methods, with CoRRPUS (specific functions) performing the best at a ROC-AUC score of 0.79.

Subjectively comparing CoRRPUS to GPT-3 shows that GPT-3 tends to struggle with the parsing of the sentences, not knowing what is relevant.
Take, for example, the sentence
``Mark Woodbury, a middle-aged man with graying hair and a mustache, smiled at Shannon as she walked into his office.''
CoRRPUS generates
\begin{itemize}
\item \texttt{self.Mark\_Woodbury.appearance.
append('graying hair')}
\item \texttt{self.Mark\_Woodbury.appearance.
append('mustache')}
\item 
\texttt{self.Mark\_Woodbury.age.
append('middle-aged')}
\end{itemize}
Meanwhile, 
GPT-3 generates ``Mark is a middle-aged man with graying hair and a mustache.''
Our particular way of prompting with CoRRPUS uses pre-specified attributes in the \texttt{character} class initialization.
By pointing out what types of attributes the model should be paying attention to (e.g. appearance or relations), CoRRPUS is better able to extract the relevant information from the natural language sentences of the given story. Meanwhile, GPT-3 ends up summarizing the original sentence.

However, among our three different CoRRPUS prompting methods, we found that their performance are similar, with the Abstract Functions performing the worst and the Specific Functions performing the best. Their similarity in score could stem from the simplicity of the type of information that needs to be reasoned over, namely attributes of characters---with no functions over verbs and how they unfold. Because of this, the Abstract Functions prompting ends up being a collection of ``set'' functions (Appendix Figure \ref{fig:re3-abstract}), which is probably making the code more complicated and giving Codex a harder time following it.

However, it's uncertain why Specific Functions perform the best on this task.
It could be due to keeping the settings of attributes better separated for each sentence via unique functions (Appendix Figure \ref{fig:re3-specific}), instead of simply separated by comments (Appendix Figure \ref{fig:re3-comment})---since comments are never operated on in real code.

\section{Conclusion}
We present CoRRPUS, a code-based prompting system that can extract structured information from complicated stories using 1-shot prompting. We conducted experiments on bAbI's Task 2 to show that code-based prompting can leverage symbolic information to perform multi-step logical reasoning better than natural-language--based prompting, regardless of whether a Code-LLM or regular LLM is used. We also evaluate CoRRPUS on detecting story inconsistencies using the Re$^3$ task \cite{yang2022re3}, showing that Code-LLMs can extract relevant information from story sentences better than LLMs. We emphasize that a careful prompting procedure that provides relevant low-level semantic abstractions can greatly improve the accuracy and generalizability of neurosymbolic reasoning but the type of prompt needed is entirely task-dependent.

\section{Limitations}
We recognize that this work was only performed on two tasks related to story understanding, thus it is difficult to say exactly how robust it really is. However, given the capabilities of LLMs and Code-LLMs, we believe our prompting techniques or similar will prove to be useful to the story understanding community.

Our work also assumes that the CoRRPUS will be asked the same question across stories. In other words, given an example as a prompt, CoRRPUS will follow that example to generate code for the next story. We are not providing the task to CoRRPUS and having it interpret the question to figure out what it should be tracking. We simply tell it to track certain information (e.g., objects, physical features of characters) so that it can solve these tasks. Therefore, for CoRRPUS to work, the user would need to know what information is salient for their task and prompt it to the system.

Even though the Re$^3$ dataset contains more complicated sentences than bAbI, these are still relatively simple English sentences. We do not know how CoRRPUS would perform on more complex stories or on stories in other languages.

Lastly, there is the issue of access. Due to cost, we were unable to rerun all of the GPT-3 experiments. The pricing of GPT-3 not only hinders new research, but it hinders reproducability efforts such as ours. Furthermore, as of the publication of this paper Codex has been removed from the OpenAI API, and it is as-of-yet unknown if GPT-3.5 or GPT-4 can handle code-based prompting as well. There are, however, still other code-based LLMs available, such as Github's Copilot and Hugging Face's Starcoder.

\section{Risks}
Working with LLMs is always a risk in itself since they are trained on huge amounts of data, some of which has never been read by developers. This text can include racist, misogynistic, queerphobic, etc. sentiments. Although the risk of harmful text might be reduced in a Code-LLM, comments, variables, and function names might still contain harmful messaging and should always be used with caution, especially when used outside of controlled research settings.

Furthermore, any code that CoRRPUS produces is not guaranteed to run nor is it guaranteed to be completely accurate in its reasoning. However, story understanding is a relatively safe testing space for reasoning and understanding tasks. 

\section*{Acknowledgements}
This material is based upon work supported by the National Science Foundation under Grant \#2030859 to the Computing Research Association for the CIFellows Project.

\bibliography{anthology,lit}
\bibliographystyle{acl_natbib}

\section{Appendix}
\subsection{Prompts}
\label{re3_examples}
These three figures illustrate the comment (Figure \ref{fig:re3-comment}), specific function (Figure \ref{fig:re3-specific}), and abstract function (Figure \ref{fig:re3-abstract}) prompts on the Re$^3$ task for the following story:
\begin{quote}
The story is set in the present day and takes place in the United States.
Joan Westfall is a woman who has died in a car accident. She is a kind and sympathetic person who is eager to help the
people she left behind.
Brent Westfall is Joan’s husband. He is a kind and loving man who is struggling to cope with his wife’s death.
Jason Westfall is Joan’s son. He is a young boy who is struggling to understand his mother’s death.
Jason Westfall sits on the floor, looking at the empty box that used to hold his sister-in-law’s belongings.
His gaze is unfocused. his dark blue eyes brimming with tears.
He cries for hours, eventually falling asleep from exhaustion.
When he wakes up, he feels dazed and ill.
Joan died in a car accident on a rainy day several weeks ago.
Jason has been carrying on with life ever since as best he can manage, but he still doesn't really know how to cope with Joan’s death.
\end{quote}

\begin{figure*}[t]
\begin{lstlisting}
## The story is set in the present day and takes place in the United States.
## Joan Westfall is a woman who has died in a car accident. She is a kind and sympathetic person who is eager to help the people she left behind.
## Brent Westfall is Joan's husband. He is a kind and loving man who is struggling to cope with his wife's death.
## Jason Westfall is Joan's son. He is a young boy who is struggling to understand his mother's death.
## Jason Westfall sits on the floor, looking at the empty box that used to hold his sister-in-law's belongings. 
## His gaze is unfocused. his dark blue eyes brimming with tears.
## He cries for hours, eventually falling asleep from exhaustion.
## When he wakes up, he feels dazed and ill.
## Joan died in a car accident on a rainy day several weeks ago.
## Jason has been carrying on with life ever since as best he can manage, but he still doesn't really know how to cope with Joan's death.
## Create a world model state to track each character's appearance, personality, and relations with other characters.

class character:
    def __init__(self, name):
        self.name = name
        self.appearance = []
        self.occupation = []
        self.gender = []
        self.age = []
        self.relations = {}
        
class World:
    def __init__(self):
        self.Joan_Westfall = character('Joan Westfall')
        self.Jason_Westfall = character('Jason Westfall')
        self.Brent_Westfall = character('Brent Westfall')
        
    def story(self):
        ## The story is set in the present day and takes place in the United States.
        ## Joan Westfall is a woman who has died in a car accident. 
        ## She is a kind and sympathetic person who is eager to help the people she left behind.
        self.Joan_Westfall.gender.append('female')
        ## Brent Westfall is Joan's husband. He is a kind and loving man who is struggling to cope with his wife's death.
        self.Joan_Westfall.relations['husband'] = 'Brent_Westfall'
        self.Brent_Westfall.relations['wife'] = 'Joan_Westfall'
        self.Brent_Westfall.gender.append('male')
        ## Jason Westfall is Joan's son. He is a young boy who is struggling to understand his mother's death.
        self.Joan_Westfall.relations['son'] = 'Jason_Westfall'
        self.Jason_Westfall.relations['mother'] = 'Joan_Westfall'
        self.Jason_Westfall.age.append('young')
        self.Jason_Westfall.gender.append('male')
        ## Jason Westfall sits on the floor, looking at the empty box that used to hold his sister-in-law's belongings. 
        self.Jason_Westfall.relations['sister_in_laws'] = 'Joan_Westfall'
        ## His gaze is unfocused. his dark blue eyes brimming with tears.
        self.Jason_Westfall.appearance.append("dark blue eyes")
        ## He cries for hours, eventually falling asleep from exhaustion.
        ## When he wakes up, he feels dazed and ill.
        ## Joan died in a car accident on a rainy day several weeks ago.
        ## Jason has been carrying on with life ever since as best he can manage, but he still doesn't really know how to cope with Joan's death.
\end{lstlisting}
\caption{Prompt using CoRRPUS (comment) on Re$^3$}
\label{fig:re3-comment}
\end{figure*}

\begin{figure*}[t]
\begin{lstlisting}
### Create a world model state and track each character's appearance, personality, relationship to other characters, and other cruical attributes.
class character:
    def __init__(self, name):
        self.name = name
        self.appearance = []
        self.occupation = []
        self.gender = []
        self.age = []
        self.relations = {}
        
class World:
    def __init__(self):
        self.Joan_Westfall = character('Joan Westfall')
        self.Jason_Westfall = character('Jason Westfall')
        self.Brent_Westfall = character('Brent Westfall')
        
    def story(self):
        self.the_story_is_set_in_the_present_day_and_takes_place_in_the_united_states()
        self.joan_westfall_is_a_woman_who_has_died_in_a_car_accident()
        self.she_is_a_kind_and_sympathetic_person_who_is_eager_to_help_the_people_she_left_behind()
        self.brent_westfall_is_joans_husband_he_is_a_kind_and_loving_man_who_is_struggling_to_cope_with_his_wife_s_death()
        self.jason_westfall_is_joans_son_he_is_a_young_boy_who_is_struggling_to_understand_his_mother_s_death()        
        self.jason_westfall_sits_on_the_floor_looking_at_the_empty_box_that_used_to_hold_his_sister_in_laws_belongings()
        self.his_gaze_is_unfocused_his_dark_blue_eyes_brimming_with_tears()
        self.he_cries_for_hours_eventually_falling_asleep_from_exhaustion()
        self.when_he_wakes_up_he_feels_dazed_and_ill()
        self.joan_died_in_a_car_accident_on_a_rainy_day_several_weeks_ago()
        self.jason_has_been_carrying_on_with_life_ever_since_as_best_he_can_manage()
        self.but_he_still_doesnt_really_know_how_to_cope_with_joans_death()
    
    def the_story_is_set_in_the_present_day_and_takes_place_in_the_united_states(self):
        pass
        
    def joan_westfall_is_a_woman_who_has_died_in_a_car_accident(self):
        pass
    def she_is_a_kind_and_sympathetic_person_who_is_eager_to_help_the_people_she_left_behind(self):
        self.Joan_Westfall.gender.append('female')
        
    def brent_westfall_is_joan_s_husband_he_is_a_kind_and_loving_man_who_is_struggling_to_cope_with_his_wife_s_death(self):
        self.Joan_Westfall.relations['husband'] = 'Brent_Westfall'
        self.Brent_Westfall.relations['wife'] = 'Joan_Westfall'
        self.Brent_Westfall.gender.append('male')
    
    def jason_westfall_is_joan_s_son_he_is_a_young_boy_who_is_struggling_to_understand_his_mother_s_death(self):
        self.Joan_Westfall.relations['son'] = 'Jason_Westfall'
        self.Jason_Westfall.relations['mother'] = 'Joan_Westfall'
        self.Jason_Westfall.age.append('young')
        self.Jason_Westfall.gender.append('male')
    
    def jason_westfall_sits_on_the_floor_looking_at_the_empty_box_that_used_to_hold_his_sister_in_laws_belongings(self):
        self.Jason_Westfall.relations['sister_in_laws'] = 'Joan_Westfall'
        
    def his_gaze_is_unfocused_his_dark_blue_eyes_brimming_with_tears(self):
        self.Jason_Westfall.appearance.append("dark blue eyes")
        
    def he_cries_for_hours_eventually_falling_asleep_from_exhaustion(self):
        pass
        
    def when_he_wakes_up_he_feels_dazed_and_ill(self):
        pass
        
    def joan_died_in_a_car_accident_on_a_rainy_day_several_weeks_ago(self):
        pass
        
    def jason_has_been_carrying_on_with_life_ever_since_as_best_he_can_manage(self):
        pass
    def but_he_still_doesnt_really_know_how_to_cope_with_joans_death(self):
        pass
\end{lstlisting}
\caption{Prompt using CoRRPUS (specific function) on Re$^3$}
\label{fig:re3-specific}
\end{figure*}
\begin{figure*}[t]
\begin{lstlisting}
## The story is set in the present day and takes place in the United States.
## Joan Westfall is a woman who has died in a car accident. She is a kind and sympathetic person who is eager to help the people she left behind.
## Brent Westfall is Joan's husband. He is a kind and loving man who is struggling to cope with his wife's death.
## Jason Westfall is Joan's son. He is a young boy who is struggling to understand his mother's death.
## Jason Westfall sits on the floor, looking at the empty box that used to hold his sister-in-law's belongings. 
## His gaze is unfocused. his dark blue eyes brimming with tears.
## He cries for hours, eventually falling asleep from exhaustion.
## When he wakes up, he feels dazed and ill.
## Joan died in a car accident on a rainy day several weeks ago.
## Jason has been carrying on with life ever since as best he can manage, but he still doesn't really know how to cope with Joan's death.
## Create a world model state to track each character's appearance, personality, and relations with other characters.

class character:
    def __init__(self, name):
        self.name = name
        self.appearance = []
        self.occupation = []
        self.gender = []
        self.age = []
        self.relations = {}
        
class World:
    def __init__(self):
        self.Joan_Westfall = character('Joan Westfall')
        self.Jason_Westfall = character('Jason Westfall')
        self.Brent_Westfall = character('Brent Westfall')
    
    def set_appearance(self, character, appearance):
        character.appearance.append(appearance)
    
    def set_occupation(self, character, occupation):
        character.occupation.append(occupation)
    
    def set_gender(self, character, gender):
        character.gender.append(gender)
    
    def set_age(self, character, age):
        character.age.append(age)
    
    def set_relation(self, character, relation, other_character):
        character.relations[relation] = other_character.name
        
    def story(self):
        ## The story is set in the present day and takes place in the United States.
        ## Joan Westfall is a woman who has died in a car accident. She is a kind and sympathetic person who is eager to help the people she left behind.
        self.set_gender(self.Joan_Westfall, "female")
        ## Brent Westfall is Joan's husband. He is a kind and loving man who is struggling to cope with his wife's death.
        self.set_relation(self.Joan_Westfall, 'husband', self.Brent_Westfall)
        self.set_relation(self.Brent_Westfall, 'wife', self.Joan_Westfall)
        self.set_gender(self.Brent_Westfall, "male")
        ## Jason Westfall is Joan's son. He is a young boy who is struggling to understand his mother's death.
        self.set_relation(self.Joan_Westfall, 'son', self.Jason_Westfall)
        self.set_relation(self.Jason_Westfall, 'mother', self.Joan_Westfall)
        self.set_age(self.Jason_Westfall, "young")
        self.set_gender(self.Jason_Westfall, "male")
        ## Jason Westfall sits on the floor, looking at the empty box that used to hold his sister-in-law's belongings. 
        self.set_relation(self.Jason_Westfall, 'sister_in_laws', self.Joan_Westfall)
        self.set_relation(self.Joan_Westfall, 'brother_in_laws', self.Jason_Westfall)
        ## His gaze is unfocused. his dark blue eyes brimming with tears.
        self.set_appearance(self.Jason_Westfall, "dark blue eyes") 
        ## He cries for hours, eventually falling asleep from exhaustion.
        ## When he wakes up, he feels dazed and ill.
        ## Joan died in a car accident on a rainy day several weeks ago.
        ## Jason has been carrying on with life ever since as best he can manage, but he still doesn't really know how to cope with Joan's death.

\end{lstlisting}
\caption{Prompt using CoRRPUS (abstract) on Re$^3$}
\label{fig:re3-abstract}
\end{figure*}
\end{document}